\documentclass[11pt]{article}

\usepackage[final]{acl}

\usepackage{times}
\usepackage{latexsym}

\usepackage[T1]{fontenc}

\usepackage[utf8]{inputenc}

\usepackage{microtype}

\usepackage{inconsolata}

\usepackage{graphicx}
\usepackage{fontawesome}

\usepackage{hyperref}
\usepackage{url}

\usepackage{algorithm}
\usepackage{algpseudocode}
\usepackage{subcaption}

\usepackage{multirow}
\usepackage{booktabs} 
\usepackage{amsmath} 
\usepackage{amssymb}
\usepackage{adjustbox}
\usepackage{enumitem}

\usepackage{minted}
\usepackage{xcolor}    
 
\usepackage{xcolor}
\definecolor{green}{HTML}{33A02C}
\definecolor{red}{HTML}{E31A1C}

\usepackage{hyperref}
\usepackage{xcolor}
\hypersetup{
    colorlinks=true,
    linkcolor=darkblue,
    urlcolor=darkblue,
}

\usepackage{listings}
\title{DataArc-SynData-Toolkit: A Unified Closed-Loop Framework for Multi-Path, Multimodal, and Multilingual Data Synthesis}

\author{
 \textbf{Zhichao Shi\textsuperscript{1,2,3,4}}\thanks{Both authors contributed equally to this research.},
 \textbf{Cehao Yang\textsuperscript{1,2,5}}\footnotemark[1],
 \textbf{Hao Zhou\textsuperscript{1,2}}\footnotemark[1],
 \textbf{Xiaojun Wu\textsuperscript{1,2,5}},
\\
 \textbf{Huajie Li\textsuperscript{1,2}},
 \textbf{Xuhui Jiang\textsuperscript{1,2}\thanks{Correspondence: jiangxuhui@dataarctech.com; xuchengjin@dataarctech.com}},
 \textbf{Chengjin Xu\textsuperscript{1,2}\footnotemark[2]},
 \textbf{Yuanzhuo Wang\textsuperscript{4}}
 \textbf{Jian Guo \textsuperscript{2}}
\\
 \textsuperscript{1}DataArc Tech Ltd.,
 \textsuperscript{2}IDEA Research, International Digital Economy Academy,
\\
 \textsuperscript{3}School of Advanced Interdisciplinary Sciences, UCAS,
 \textsuperscript{4}Institute of Computing Technology, CAS,
\\
 \textsuperscript{5}The Hong Kong University of Science and Technology (Guangzhou)
\\
{\fontsize{14}{16}\faGithub}~\url{https://github.com/DataArcTech/DataArc-SynData-Toolkit}
}

\begin{document}
\maketitle

\begin{abstract}
Synthetic data has emerged as a crucial solution to the data scarcity bottleneck in large language models (LLMs), particularly for specialized domains and low-resource languages. However, the broader adoption of existing synthetic data tools is severely hindered by convoluted workflows, fragmented data standards, and limited scalability across modalities.
To address these limitations, we develop DataArc-SynData-Toolkit, an open-source framework featuring: (1) a configuration-driven, end-to-end pipeline equipped with an intuitive visual interface and simplified CLI for exceptional usability; (2) a unified, quality-controllable synthesis paradigm that standardizes multi-source data generation to ensure high reusability; and (3) a highly modular architecture designed for seamless multimodal, multilingual, and multi-task adaptation.
We apply the toolkit in multiple application scenarios. 
Experimental results demonstrate that our toolkit achieves an optimal balance between generation efficiency and data quality. By offering an end-to-end and visually interactive pipeline, DataArc-SynData-Toolkit significantly lowers the technical barrier to synthetic data generation and subsequent model training, accelerating its practical deployment in real-world applications.
\end{abstract}
\section{Introduction}
\label{sec: 01_intro}

The rapid evolution of Large Language Models (LLMs) has exponentially driven the demand for high-quality training corpora \citep{openai2023gpt, touvron2023llama}.
However, acquiring real-world data at scale remains a formidable challenge, particularly for specialized domains and low-resource languages \citep{barati2025searchinstruct, abdalla2025future}. Consequently, synthetic data generation has become a critical paradigm for expanding training corpora.
While several pioneering frameworks like Synthetic Data-RL \citep{guo2025synthetic} and DataFlow \citep{liang2025dataflow} have significantly advanced the field, their primary design often leans towards algorithmic exploration rather than out-of-the-box, end-to-end engineering usability.

\begin{table*}[htbp]
\centering
\resizebox{\textwidth}{!}{
\begin{tabular}{ll c ccc ccc cccc}
    \toprule
    \multirow{2}{*}{\textbf{Framework}} & \multirow{2}{*}{\textbf{Github Repository}} & \multirow{2}{*}{\textbf{\shortstack{Data\\Prep.}}} & \multicolumn{3}{c}{\textbf{Modality Support}} & \multicolumn{3}{c}{\textbf{Synthesis Strategies}} & \multirow{2}{*}{\textbf{Filter}} & \multirow{2}{*}{\textbf{Eval.}} & \multirow{2}{*}{\textbf{\shortstack{Model-\\Training}}} & \multirow{2}{*}{\textbf{Vis.}}\\
    
    \cmidrule(lr){4-6} \cmidrule(lr){7-9}
    & & & Struct. & Text & Multi. & Local & Distill & Web & & & & \\
    \midrule
    \textbf{DataArc (Ours)} & \href{https://github.com/DataArcTech/DataArc-SynData-Toolkit}{\color{darkblue} DataArcTech/DataArc-SynData-Toolkit} & \checkmark & \checkmark & \checkmark & \checkmark & \checkmark & \checkmark & \checkmark & \checkmark & \checkmark & \checkmark & \checkmark \\
    
    \textbf{SDV} & \href{https://github.com/sdv-dev/SDV}{\color{darkblue} sdv-dev/SDV} & \checkmark & \checkmark &  &  & \checkmark &  &  &  & \checkmark &  & \\
    
    \textbf{Gretel} & \href{https://github.com/gretelai/gretel-synthetics}{\color{darkblue} gretelai/gretel-synthetics} & \checkmark & \checkmark & \checkmark & & \checkmark & & & & & & \\
    
    \textbf{YData} & \href{https://github.com/ydataai/ydata-synthetic}{\color{darkblue} ydataai/ydata-synthetic} & \checkmark & \checkmark &  &  & \checkmark &  &  &  & \checkmark & & \checkmark \\
    
    \textbf{EasyInstruct} & \href{https://github.com/zjunlp/EasyInstruct}{\color{darkblue} zjunlp/EasyInstruct} &  &  & \checkmark &  & \checkmark & & & \checkmark & \checkmark & & \checkmark \\
    
    \textbf{EasyDataset} & \href{https://github.com/ConardLi/easy-dataset}{\color{darkblue} ConardLi/easy-dataset} & \checkmark & \checkmark & \checkmark & \checkmark & \checkmark & \checkmark &  & \checkmark & \checkmark & & \checkmark \\

    \textbf{EasyDistill} & \href{https://github.com/modelscope/easydistill}{\color{darkblue} modelscope/easydistill} & \checkmark & & \checkmark & & & \checkmark & & \checkmark & \checkmark & \checkmark & \\
    
    \textbf{DataFlow} & \href{https://github.com/OpenDCAI/DataFlow}{\color{darkblue} OpenDCAI/DataFlow} & \checkmark & \checkmark & \checkmark &  & \checkmark & & & \checkmark & \checkmark & \checkmark & \checkmark \\

    \textbf{Distilabel} & \href{https://github.com/argilla-io/distilabel}{\color{darkblue} argilla-io/distilabel} & \checkmark & \checkmark & \checkmark & & \checkmark & \checkmark & \checkmark & \checkmark & \checkmark & \checkmark & \\

    \textbf{SyntheticDataKit} & \href{https://github.com/meta-llama/synthetic-data-kit}{\color{darkblue} meta-llama/synthetic-data-kit} & \checkmark & \checkmark & \checkmark & \checkmark & \checkmark & & \checkmark & \checkmark & & & \\

    \textbf{Synthetic Data RL} & \href{https://github.com/gydpku/Data_Synthesis_RL}{\color{darkblue} gydpku/Data\_Synthesis\_RL} & \checkmark & & \checkmark & & \checkmark & & & \checkmark & \checkmark & \checkmark & \\
    
    \bottomrule
\end{tabular}
}
\caption{Feature comparison of DataArc with other open-source synthetic data frameworks. \textit{Data Prep.} refers to Data Preprocessing; \textit{Struct.} and \textit{Multi.} denote Structured Data and Multimodal Data, respectively. For Synthesis Strategies, ``Local'' indicates synthesis based on local corpora, ``Distill'' refers to knowledge distillation, and ``Web'' denotes synthesis via web search. As shown, DataArc is the only framework offering an end-to-end pipeline covering multiple strategies, multimodal synthesis, and downstream model fine-tuning.}
\label{tab:framework_comparison}
\end{table*}

Specifically, as shown in Table~\ref{tab:framework_comparison}, current synthetic data tools face three major challenges in practical scenarios:
(1) \textbf{Steep Learning Curves and Limited Usability}. Existing tools cater primarily to AI experts. Users lacking specialized backgrounds struggle to navigate the fragmented processes of data generation, model training, and validation, often resorting to complex, multi-tool scripting.
(2) \textbf{Fragmented Workflows and Unstandardized Pipelines}. There is a distinct lack of a unified paradigm for handling multi-source data \citep{chen2024data}. This fragmentation leads to unpredictable data quality and severely limits cross-project reusability.
(3) \textbf{Restricted Modality and Language Scalability}. Most frameworks lack generic interfaces for multimodal and low-resource language synthesis, resulting in prohibitively high extension and maintenance costs.

To address these challenges, we propose \textbf{DataArc-SynData-Toolkit}, a configuration-driven end-to-end system for data synthesis, model training, and evaluation. The toolkit unifies data collection, synthesis, quality control, post-training, and evaluation into a single closed-loop workflow that supports continuous iteration. Users can construct complex synthetic datasets and optimize models through concise configuration files. In addition, the toolkit encapsulates execution logic into simplified commands and provides an intuitive visual interface, which significantly improves usability.

The toolkit adopts a highly modular and agent-based system design. It unifies the synthesis pipelines and data formats for multi-source synthetic data, including local corpora, open-source web datasets, and model distillation. This design ensures high data quality and strong reusability. Furthermore, the system abstracts unified adaptation interfaces for multimodal and multilingual data, greatly reducing the cost of future extensions.

In summary, DataArc-SynData-Toolkit is designed with usability, standardization, and scalability at its core. Its main contributions are as follows:
\begin{itemize}[leftmargin=*]
\item We introduce a configuration-driven, closed-loop system encompassing data synthesis, model training, and evaluation, accessible via a streamlined CLI and an intuitive UI to democratize synthetic data generation.
\item We design a unified, standardized paradigm for multi-source data synthesis equipped with robust quality-control mechanisms, thereby guaranteeing data reliability and reusability.
\item We deliver a highly extensible, modular framework inherently capable of multimodal and multilingual synthesis, significantly lowering the engineering overhead for future academic and industrial adaptations.
\end{itemize}
\section{Framework}
\label{sec: 02_framework}

\begin{figure*}[ht]
    \centering
    \includegraphics[width=\linewidth]{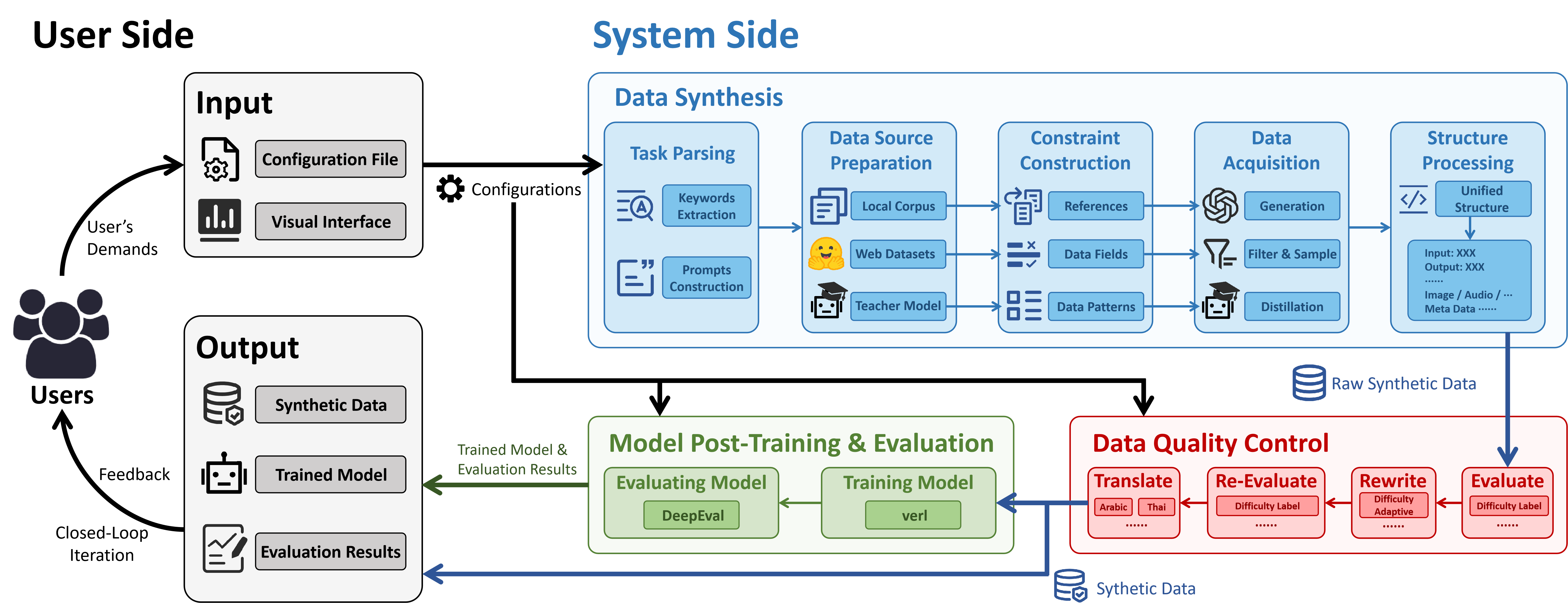}
    \caption{The overview of DataArc-SynData-Toolkit. \textbf{User side}:  Users only need to set configurations in the file or the visual interface to obtain the desired synthetic data,  models, and evaluation results. \textbf{System Side}: The toolkit divides the end-to-end pipeline into three stages: (1) Data synthesis: synthesize data based on the configuration; (2) Quality control: filter and rewrite the data to ensure quality; (3) Model post-training and evaluation.}
    \label{fig:overview}
    \vspace{-5pt}
\end{figure*}



We aim to develop a user- and developer-friendly synthetic data toolkit. 
The toolkit should provide a highly extensible framework for developers, and offer users an end-to-end pipeline and an easy-to-use operational paradigm.

Therefore, DataArc-SynData-Toolkit adopts a unified and modular agent-based design. 
We generalize and abstract the process from multi-source synthetic data to model outputs. 
Based on the convergence nodes in the pipeline, and considering the mergeability of different steps, we divide the entire end-to-end workflow into three key stages: data synthesis, data quality control, and model post-training and evaluation, as shown in Figure~\ref{fig:overview}.
In this section, we introduce the design of DataArc-SynData-Toolkit for implementing the key stages.

\subsection{Data Synthesis}
\label{subsec: 2_1_data_synthesis}

To address the lack of unified workflows and data standards in existing tools, DataArc-SynData-Toolkit abstracts a unified data synthesis paradigm, including \textit{task parsing}, \textit{data source preparation}, \textit{constraint construction}, \textit{data acquisition}, and \textit{structured processing}.
(1) \textit{Task parsing} parses the task definitions into information like task instructions and domain boundaries.
(2) \textit{Data source preparation} selects and prepares data sources based on the configuration, including local corpora, web datasets, or models to be distilled.
(3) \textit{Constraint construction} builds constraints for data synthesis, including reference passages, field limitations, or pattern constraints.
(4) \textit{Data acquisition} generates or filters the required data from the selected sources based on the parsed information and the defined constraints.
(5) \textit{Structured processing} normalizes the synthesized data into a unified format.

Based on this unified paradigm, DataArc-SynData-Toolkit designs \texttt{BaseTaskExecutor}. 
As shown in Figure~\ref{fig:code_data_synthesis}, \texttt{BaseTaskExecutor} has implemented basic execution logic. Developers can flexibly extend different synthesis paths by overriding the abstract functions.
To ensure data normalization and reusability, the toolkit processes all the data into the same format in Function \texttt{structure\_process}.

We have implemented three paths: local corpus–driven synthesis (\texttt{LocalTaskExecutor}), web dataset–based filtering (\texttt{WebTaskExecutor}), and instruction distillation–based synthesis (\texttt{DistillTaskExecutor}), which cover the common synthesis paths in current practice.


\begin{figure}[htbp]
\vspace{-\intextsep}
\begin{lstlisting}[language=python]
from sdgsystem.tasks import BaseTaskExecutor

# Inputs: configurations, llm
# Initialization
executor = BaseTaskExecutor(configurations, llm)
# Step(1): parsing configurations
parsed_information = executor.task_parsing()
# Step(2): prepare data sources
sources = executor.prepare()
# Step(3): get constraints for synthesis
constraints = executor.construct_constraints(parsed_information)
# Step(4): synthesizing data from sources
raw_data = executor.data_acquisition(
    sources, 
    constraints, 
    parsed_information
)
# Step(5): processing data into a unified structure
syn_data = executor.structure_process(raw_data)
# Outputs: synthetic data in format: 
# {input: ..., output: ..., image: ..., audio: ..., metadata: ...}
\end{lstlisting}
\caption{The code implementation for data synthesis in abstract class \texttt{BaseTaskExecutor}}
\label{fig:code_data_synthesis}
\vspace{-1em}
\end{figure}

\subsubsection{Local Task}
\label{subsubsec: 2_1_1_local}
This path synthesizes data based on local corpora.
Function \texttt{task\_parsing} extracts keywords from the task instruction and domain in configurations for retrieval.
Function \texttt{prepare} loads the retriever for the local corpora and prepares the generator for synthesis.
Function \texttt{construct\_constraints} invokes the retriever to retrieve reference passages from local corpora using methods such as BM25\citep{robertson2009probabilistic} or dense methods\citep{karpukhin2020dense, khattab2020colbert}.
Function \texttt{data\_acquisition} combines the task instruction, reference passages, format constraints, and optional examples to form a prompt. The LLM generator is prompted to synthesize data.

\subsubsection{Web Task}
\label{subsubsec: 2_1_2_web}
This path selects required data from open-source web datasets.
Function \texttt{task\_parsing} is the same as Local Task.
Function \texttt{prepare} searches candidate datasets from open platforms (e.g., HuggingFace) based on the extracted keywords.
Function \texttt{construct\_constraints} adopts the LLM to select the data fields that best match the task based on configurations.
Function \texttt{data\_acquisition} extracts the contents in the selected fields, and scores the dataset by the task consistency and quality of these contents. The final data are preferentially sampled from datasets with higher scores until the required data volume is reached.

\subsubsection{Distill Task}
\label{subsubsec: 2_1_3_distill}
This path distills data from a stronger teacher model based on the task instruction.
Function \texttt{task\_parsing} parses the configurations to get parameters for the teacher model.
Function \texttt{prepare} deploys the teacher model.
Function \texttt{construct\_constraints} queries the teacher model to extract high-quality generation patterns from task instruction and optional examples.
Function \texttt{data\_acquisition} prompts the teacher model with task instruction, pattern constraints, and format constraints to generate data.

\subsection{Data Quality Control}
\label{subsec: 2_2_quality_control}
Caused by factors such as inappropriate sample difficulty\citep{tong2024dart, ferracci2024targeted}, directly training models on initial synthetic data may lead to degraded performance. 
Therefore, DataArc-SynData-Toolkit designs a quality control stage to ensure the data quality.
This stage is divided into two key functions: evaluation and rewrite.
Through evaluation, rewriting, and re-evaluation, high-quality synthetic data is generated.


To support extensibility across different algorithms, we highly abstract the evaluation and rewriting steps into base classes \texttt{Evaluator} and \texttt{BaseRewriter}. 
The evaluator takes raw synthetic data as input and outputs corresponding evaluation scores. 
The rewriter takes the raw data and their scores as input and produces rewritten samples based on these scores. 
Developers can override the abstract functions and design different strategies, enabling flexible extension of quality control policies.
An example of the code implementation for quality control in the toolkit is provided below.


\begin{figure}[htbp]
\vspace{-\intextsep}
\begin{lstlisting}[language=python]
from sdgsystem.evaluation import Evaluator
from sdgsystem.generation.rewriter import DifficultyAdjustRewriter

# Inputs: configs, llm, base_model to be trained.
evaluator = Evaluator(configs, base_model, llm)
rewriter = DifficultyAdjustRewriter(configs, llm)

# evaluation
init_eval = evaluator.evaluate(dataset)
# rewrite
rewritten_dataset = rewriter.rewrite(
    dataset, 
    init_eval,
)
# re-evaluation
final_eval = evaluator.evaluate(
    rewritten_dataset
)
solved, learnable, unsolved \
= rewritten_dataset.categorize_by_score(final_eval)
# learnable subset is selected as training data
\end{lstlisting}
\caption{An example of code implementation for quality control strategy in the toolkit.}
\label{fig:code_quality_control}
\vspace{-5pt}
\end{figure}

\subsection{Model Post-Training and Evaluation}
\label{subsec: 2_3_model_training}


To facilitate users in examining the synthetic data's training effectiveness, DataArc-SynData-Toolkit integrates model post-training and evaluation modules, establishing a closed-loop model iteration.

For training, the toolkit integrates a training module powered by verl\citep{verl}, enabling users to train models directly on their synthesized data. 
The toolkit supports two training methods: Supervised Fine-tuning (SFT) and Group Relative Policy Optimization (GRPO)\citep{grpo}.

For evaluation, the toolkit provides a model evaluation module powered by DeepEval\footnote{\url{https://github.com/confident-ai/deepeval}}. 
This module enables users to evaluate post-trained models on user-provided evaluation datasets and applies G-Eval\citep{geval} to judge model responses. 
The toolkit supports three metrics: Answer Correctness, Format Compliance, and Pairwise Preference.

\subsection{Other Extensible Modules}
\label{subsec: 2_4_extensible_module}


To make it easy for developers to integrate various algorithms and strategies into the toolkit, we adopt a highly modular design. 
In the previous subsections, we described the agent-based abstraction and encapsulation of the three key stages of the workflow. 
In this section, we introduce other extensible modules, with a focus on multimodal and multilingual adaptation modules.

For the multimodal adapter, we developed \texttt{MMProjector}, which executes the task of synthesizing multimodal instructions after being provided with seed image data for cold-start initialization.
The base model and pipeline-driven model are switched to their multimodal versions; consequently, the final synthesized instructions include not only \texttt{input} and \texttt{output}, but also the corresponding seed images. 
This component can be optionally activated when users provide seed images or when processing local rich-text data.

For the multilingual adapter, we design \texttt{BaseTranslator}, which performs language conversion in the final step of quality control.
The toolkit currently integrates adapters for low-resource languages such as Arabic.

In addition, we abstract the execution logic for sequential data and design \texttt{ParallelExecutor}. 
This executor encapsulates parallelization, temporary interruption, data caching, and resumption from existing progress. 
Developers can call Function \texttt{execute(...)} of \texttt{ParallelExecutor} to perform more efficient and reliable processing of sequential data, as shown in Figure~\ref{fig:parallel_executor}.

\begin{figure}[htbp]
\vspace{-\intextsep}
\begin{lstlisting}[language=python]
from sdgsystem.parallel import ParallelExecutor

executor = ParallelExecutor(n_workers=10)
# define the function to process each item
def process(...): ...
# Inputs: the iterable sequence data, the process function
outputs = executor.execute(
    iterable_inputs=data, 
    process_function=process, 
    ...
)
# outputs: A list like [r1, r2, ..., None, ...]
# None indicates items failed to be processed when the program is interrupted.
\end{lstlisting}
\caption{An example of \texttt{ParallelExecutor}.}
\label{fig:parallel_executor}
\end{figure}

\subsection{Simplified Usage}
\label{subsec: 2_5_usage}


DataArc-SynData-Toolkit is designed with usability as the primary goal, enabling even users without professional backgrounds to easily synthesize data.
Therefore, the toolkit provides highly simplified commands for CLI and an intuitive visual interface to launch the workflow.

The toolkit offers simplified commands to start the pipeline. 
We use uv\footnote{\url{https://github.com/astral-sh/uv}} as the package manager and unify the command format. 
Users can complete the whole pipeline with simple commands: 

\begin{figure}[htbp]
\vspace{-\intextsep}
\begin{lstlisting}
uv run sdg generate configs/sdg.yaml
uv run sdg train configs/[sft|grpo].yaml
uv run sdg eval configs/eval.yaml
\end{lstlisting}
\caption{Simplified commands for quick start in CLI.}
\label{fig:code_quickstart}
\end{figure}

In addition, we provide a visual interface to facilitate task configuration, data management, and model management for users. 
The toolkit adopts a frontend–backend separated architecture, featuring a FastAPI backend and a React frontend for improved visualization and scalability.
The detailed design is described in Appendix~\ref{appendix: detail_interface_design}.

The samples of data synthesized by our toolkit are detailed in Appendix~\ref{appendix: sample}. 
Detailed deployment instructions, command usage, and tutorial videos for the visual interface are available in our GitHub repository\footnote{\url{https://github.com/DataArcTech/DataArc-SynData-Toolkit}}. 
\section{Performance Evaluation}
\label{sec: 03_practice}


In this section, we systematically evaluate the performance of DataArc-SynData-Toolkit, including the performance across different domains, multi-lingual, multi-modal, and its efficiency.

\subsection{Benchmarks}
\label{subsec: 3_1_benchmarks}

We conduct experiments on domain-specific, multi-lingual, and multi-modal benchmarks. 

\begin{itemize}[leftmargin=*] 
  \item \textbf{Domain-specific benchmarks:} \textbf{MedQA} \citep{medqa} contains multiple-choice questions collected from medical board exams in the US and China. We adopt the subset MedQA-USMLE-4-options \footnote{\url{https://huggingface.co/datasets/GBaker/MedQA-USMLE-4-options}} of 1,273 questions. \textbf{Flare-CFA} \footnote{\url{https://huggingface.co/datasets/TheFinAI/flare-cfa}} comprises 1,032 multiple-choice questions covering CFA exam levels I and II. \textbf{LexEval} \citep{li2024lexeval} evaluates LLM performance in legal applications, and 500 multiple-choice questions involving multi-hop reasoning are adopted.
  \item \textbf{Multi-lingual benchmark:} \textbf{Arabic Broad Benchmark (ABB)}\footnote{\url{https://huggingface.co/datasets/silma-ai/arabic-broad-benchmark}} is a human-validated and compact benchmark of 470 free-form questions spanning 22 Arabic language tasks \citep{ABBL}.
  \item \textbf{Multi-modal benchmarks:} \textbf{FinMME} \citep{finmme} contains high-quality financial research samples across 18 domains, and about 1,850 numerical questions are selected. \textbf{TableVQA} \citep{TableVQA} contains table question-answering and table structure recognition datasets of 1,500 samples.
\end{itemize}

\begin{table*}[ht]
\centering
\begin{tabular}{lccccccc}
\toprule
\multirow{2}{*}{\textbf{Method}} &
\multirow{2}{*}{\textbf{Data Size}} &
\multicolumn{3}{c}{\textbf{Domain}} &
\multicolumn{1}{c}{\textbf{Multi-lingual}} &
\multicolumn{2}{c}{\textbf{Multi-modal}} \\
\cmidrule(lr){3-5}\cmidrule(lr){6-6}\cmidrule(lr){7-8}
& &
\textbf{MedQA} & \textbf{Flare-CFA} & \textbf{LexEval} &
\textbf{ABB} & 
\textbf{FinMME} & \textbf{MedQA} \\
\midrule

\multicolumn{2}{l|}{} &
\multicolumn{4}{c|}{\textit{Qwen2.5-7B}} &
\multicolumn{2}{c}{\textit{Qwen2.5-VL-7B}} \\
\midrule

Base Model     & - & 42.34   & 52.91 & 19.80 & 6.42 & 16.41 & 55.13 \\
         & 1000  & 59.02 & 67.15 & 33.87 & 6.96 & 24.90 & 61.90 \\
\textsc{SynData}  & 2000  & 64.57 & 73.93 & 42.80 & 7.08 & 26.98 &  61.39 \\
         & 4000  & 68.12 & 75.80 & 46.15 & 7.12 & 32.47 & 64.32 \\
\midrule

\multicolumn{2}{l|}{} &
\multicolumn{4}{c|}{\textit{Qwen3-4B}} &
\multicolumn{2}{c}{\textit{Qwen3-VL-4B}} \\
\midrule

Base Model    & -     & 57.52 & 60.89 & 31.48 & 6.26 & 27.65 & 54.10 \\
         & 1000  & 64.70 &  66.41 & 37.90 & 6.75 & 33.81 & 60.37 \\
\textsc{SynData}  & 2000  & 66.13 & 75.74 & 45.44 & 6.69 & 33.46 & 66.54 \\
         & 4000  & 70.62 & 74.41 & 48.00 & 6.77 & 36.96 & 64.90 \\
\bottomrule
\end{tabular}
\caption{The main experiment results.}
\label{tab:main_results}
\end{table*}

\subsection{Implementation Details}
\label{subsec: 3_2_implementation}

For each benchmark, we collect seed data from HuggingFace for cold start according to the corresponding task definition. 
We sample query terms from the extracted keyword set and, by default, select 5 candidate relevant datasets to ensure diversity in data distribution. 
Subsequently, we sample and filter 25 data instances and clean them into seed data based on the task definition. 
For multi-modal tasks, we will collect seed images for cold-start. 
We then use \textit{gpt-5-mini} to synthesize the initial instruction data and submit it to a quality management control pipeline powered by \textit{gpt-5-mini}. 
The resulting curated datasets are constructed at specified scales for downstream training, namely 1,000 / 2,000 / 4,000 samples. 
We validate our toolkit on a wide range of LLMs, including Qwen2.5-7B-Instruct and Qwen3-4B for domain and multi-lingual tasks, and Qwen2.5-VL-7B-Instruct and Qwen3-VL-4B-Instruct for multi-modal tasks. 
All experiments are conducted on an Ubuntu machine equipped with eight NVIDIA H800-80G GPUs.

\subsection{Evaluation Results}
\label{subsec: 3_2_eval_results}

\begin{figure}
    \centering
    \includegraphics[width=\linewidth]{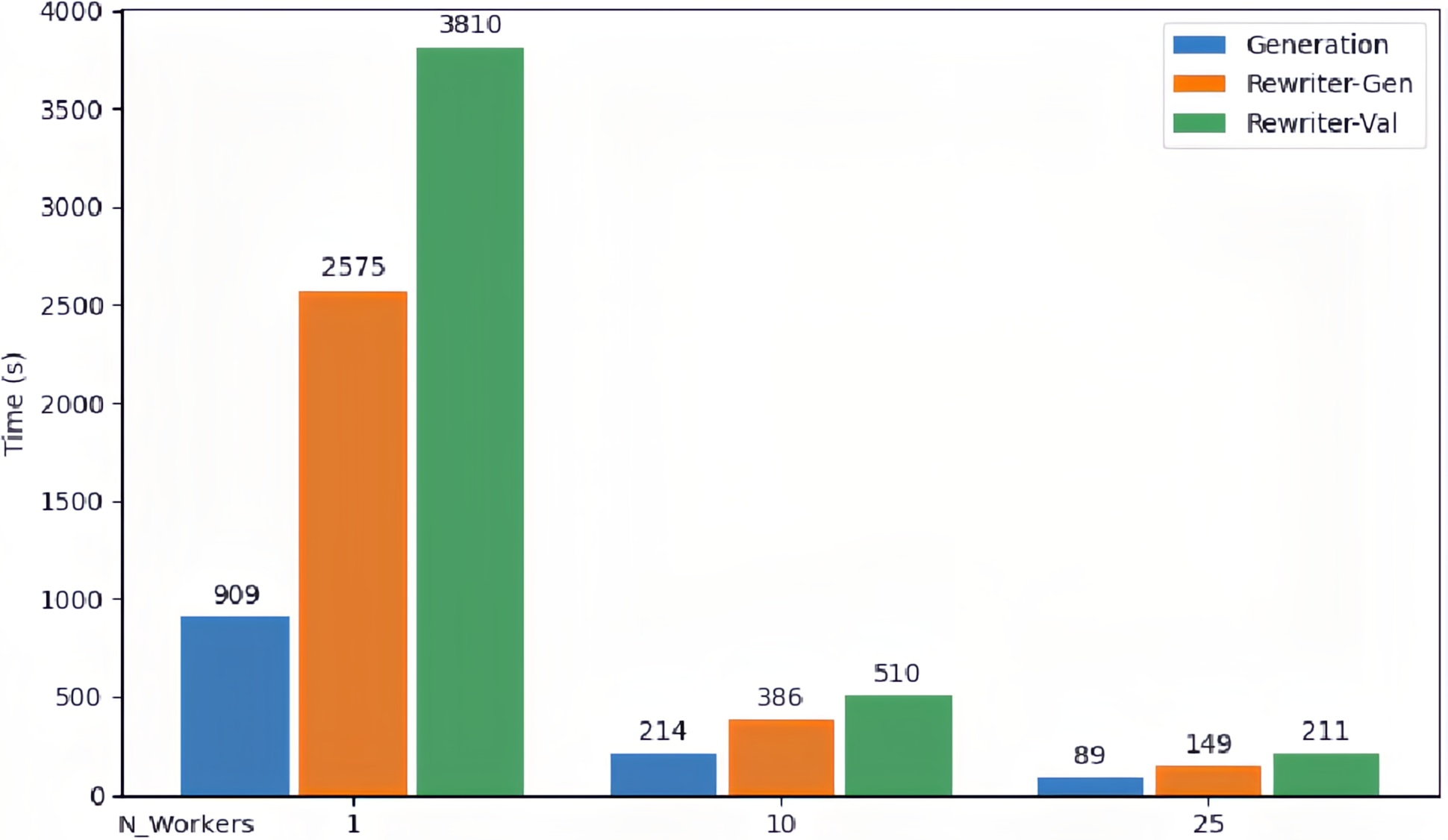}
    \caption{The efficiency of our \texttt{ParallelExecutor} design in toolkit when synthesizing 500 samples.}
    \label{fig:efficiency}
    \vspace{-5pt}
\end{figure}

\paragraph{Consistent Performance Gains Across Models.}
As shown in Table~\ref{tab:main_results}, synthetic data generated by our toolkit consistently improves performance across different models and task settings. 
For both Qwen2.5-7B and Qwen3-4B, training on synthesized samples yields substantial accuracy gains over the corresponding base models. 
For instance, Qwen2.5-7B improves from 42.34 to 68.12 on MedQA and from 19.80 to 46.15 on LexEval with 4,000 synthetic samples. 
Similar trends are observed for Qwen3-4B, which increases from 57.52 to 70.62 on MedQA and from 31.48 to 48.00 on LexEval. 
These results demonstrate that the proposed toolkit provides robust and architecture-agnostic gains, validating its effectiveness across heterogeneous model scales and modalities.

\paragraph{Scaling Synthetic Data Enhances Performance Gains.}
Larger synthetic datasets lead to more pronounced performance gains. 
Across nearly all benchmarks, accuracy steadily increases as the training data size expands from 1,000 to 2,000 and 4,000 samples. 
For example, on Qwen2.5-7B, MedQA accuracy rises from 59.02 to 64.57 and further to 68.12, while FinMME improves from 24.90 to 32.47. 
The scaling behavior suggests strong potential for further gains with increased synthetic data budgets, highlighting the practicality of synthetic data as a controllable and efficient alternative to costly human annotation.

\paragraph{Parallel Design Significantly Boosts Synthesis Efficiency.}
Experimental results in Figure~\ref{fig:efficiency} demonstrate that adjusting the number of workers in the toolkit's \texttt{ParallelExecutor} yields substantial improvements in synthesis efficiency. 
Without parallelization, generating merely 500 initial samples requires 909 seconds. As the number of active workers increases, the required time decreases markedly. Since all modules within the pipeline benefit from the parallel design, including both stages of the rewriter (i.e., generation and validation), scaling the worker count to just 10 or 25 achieves significant performance gains.










\section{Conclusion and Future Work}
\label{sec: 04_conclusion}


We propose DataArc-SynData-Toolkit, a configuration-driven, end-to-end open-source toolkit for data synthesis. 
The toolkit provides a highly extensible, unified, and quality-controllable framework for multi-path, multimodal, multilingual data synthesis. 
For the next release, we plan to enable encrypted synthetic data generation to protect sensitive information for secure model training. 
Future work will focus on incorporating more algorithms, modalities, and languages to offer broader options.

\section*{Limitations}

Due to the rapid evolution of synthetic data methods, our toolkit currently integrates only a subset of data synthesis and validation approaches, such as the difficulty-based data rewriting method shown in Figure~\ref{fig:code_quality_control}. 
However, our analysis of the data synthesis pipeline enables the toolkit to provide a highly abstract, extensible, and unified framework, with careful system optimizations. 
New methods can be easily integrated into the toolkit, which provides users with more convenient options.


\bibliography{custom}

\appendix
\section{Detailed Interface Design}
\label{appendix: detail_interface_design}

\begin{figure}[h]
    \centering
    \includegraphics[width=\linewidth]{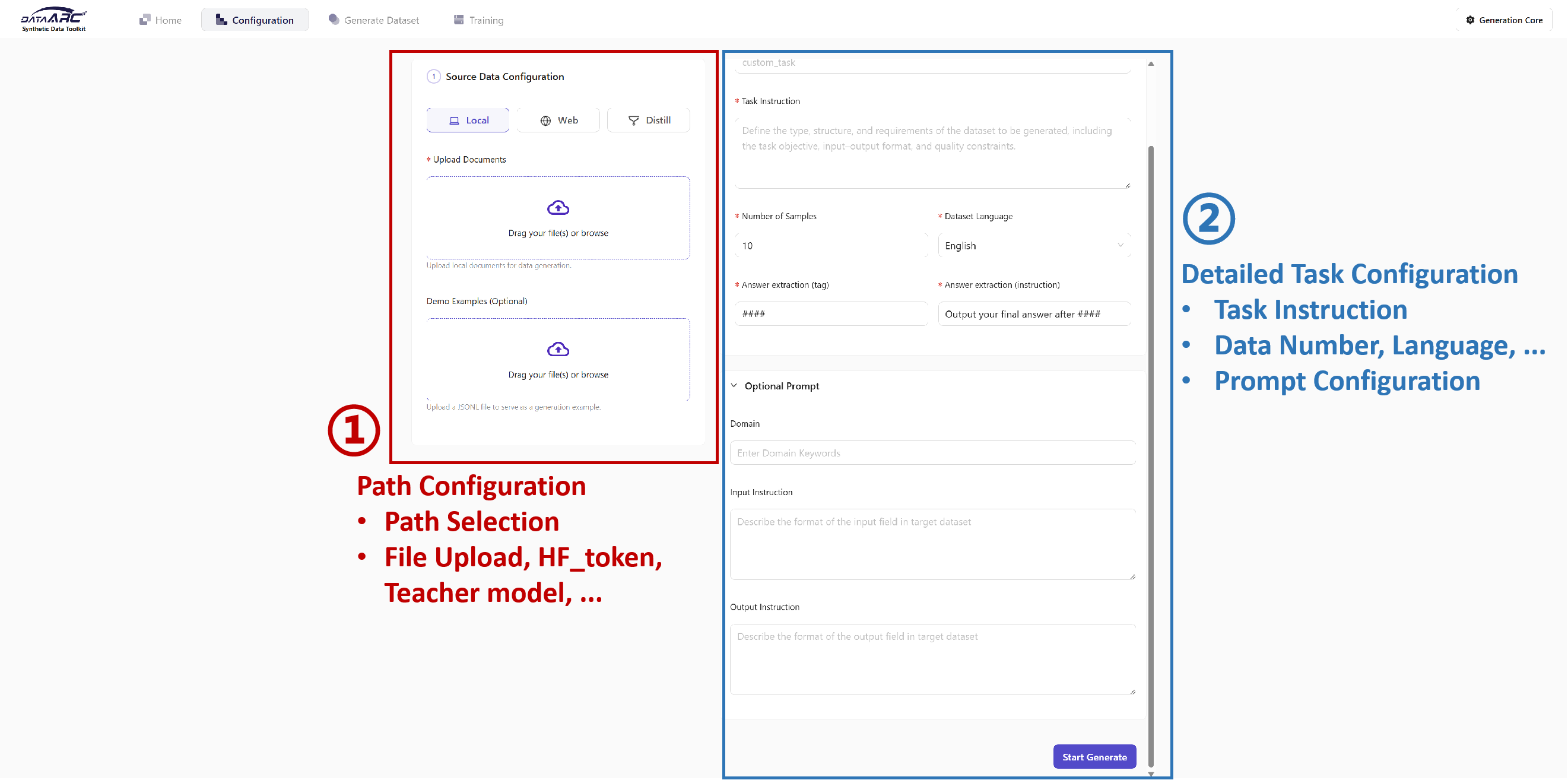}
    \caption{Interface of synthetic task configurations.}
    \label{fig:i1_task_config}
\end{figure}

\begin{figure}[h]
    \centering
    \includegraphics[width=\linewidth]{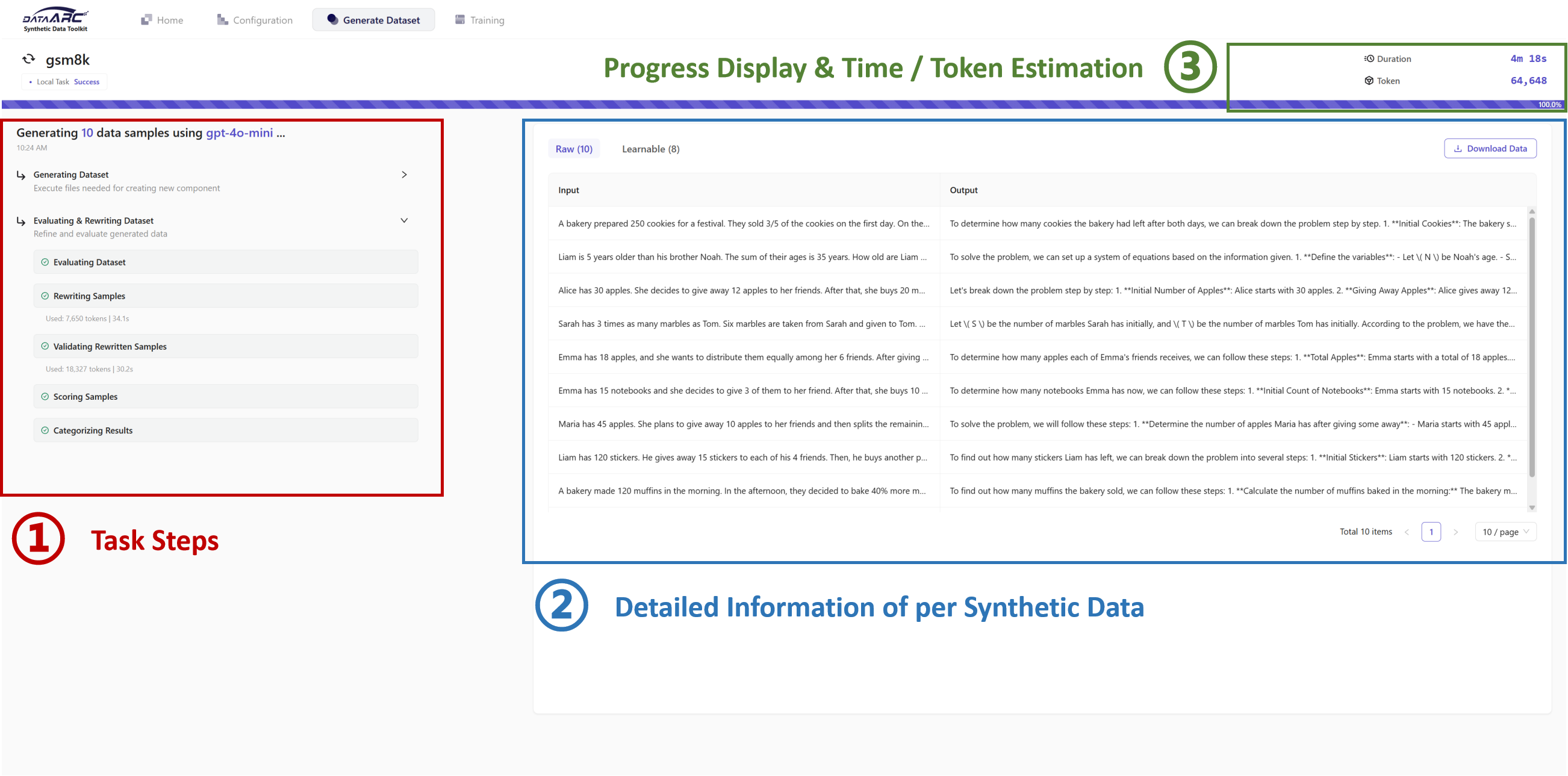}
    \caption{Visualization of data synthesis workflow.}
    \label{fig:i2_data_synthesis}
\end{figure}

\begin{figure}[h]
    \centering
    \includegraphics[width=\linewidth]{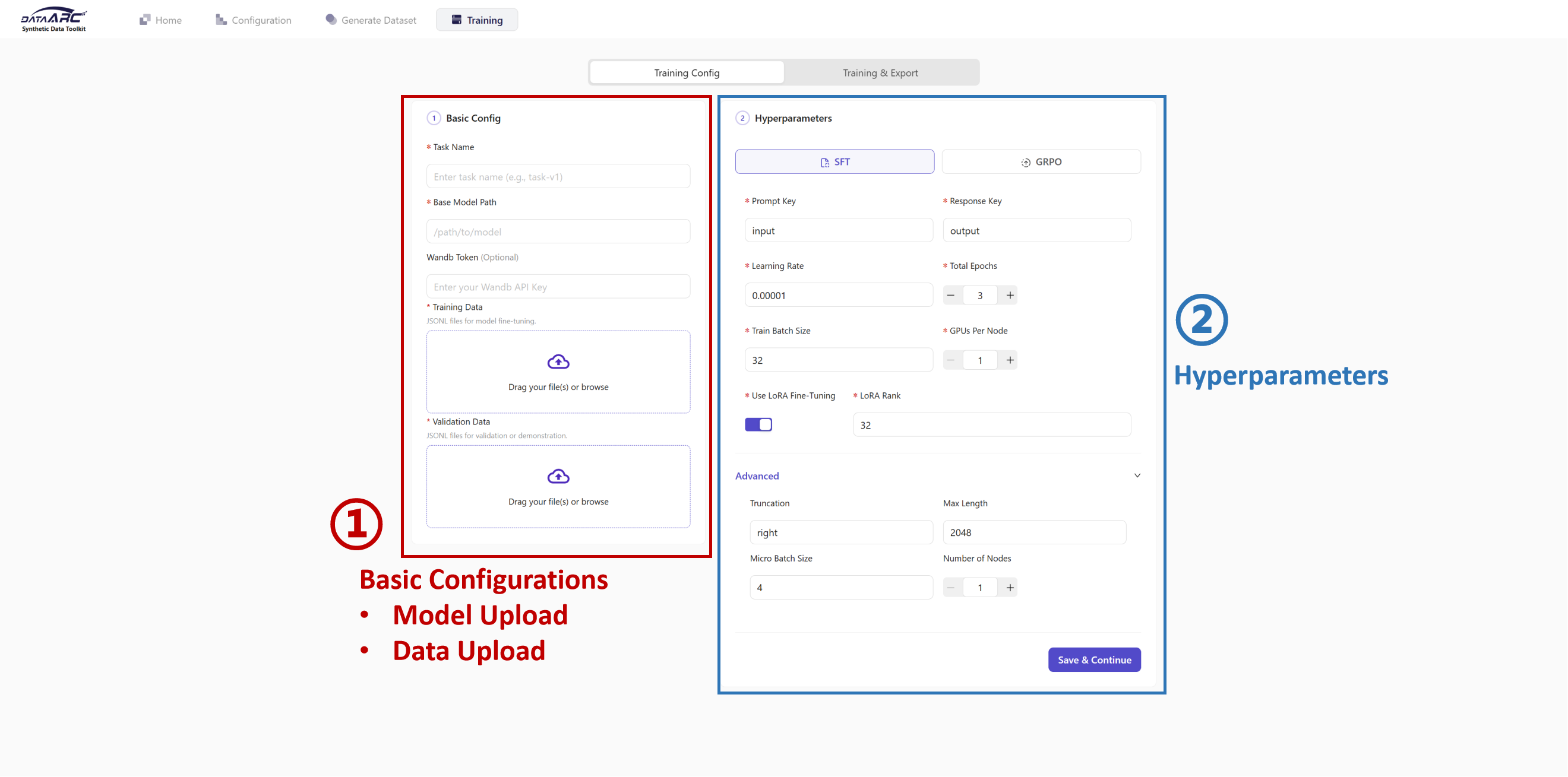}
    \caption{Interface of model training.}
    \label{fig:i3_train_config}
    \vspace{-1em}
\end{figure}

\begin{figure}[h]
    \centering
    \includegraphics[width=\linewidth]{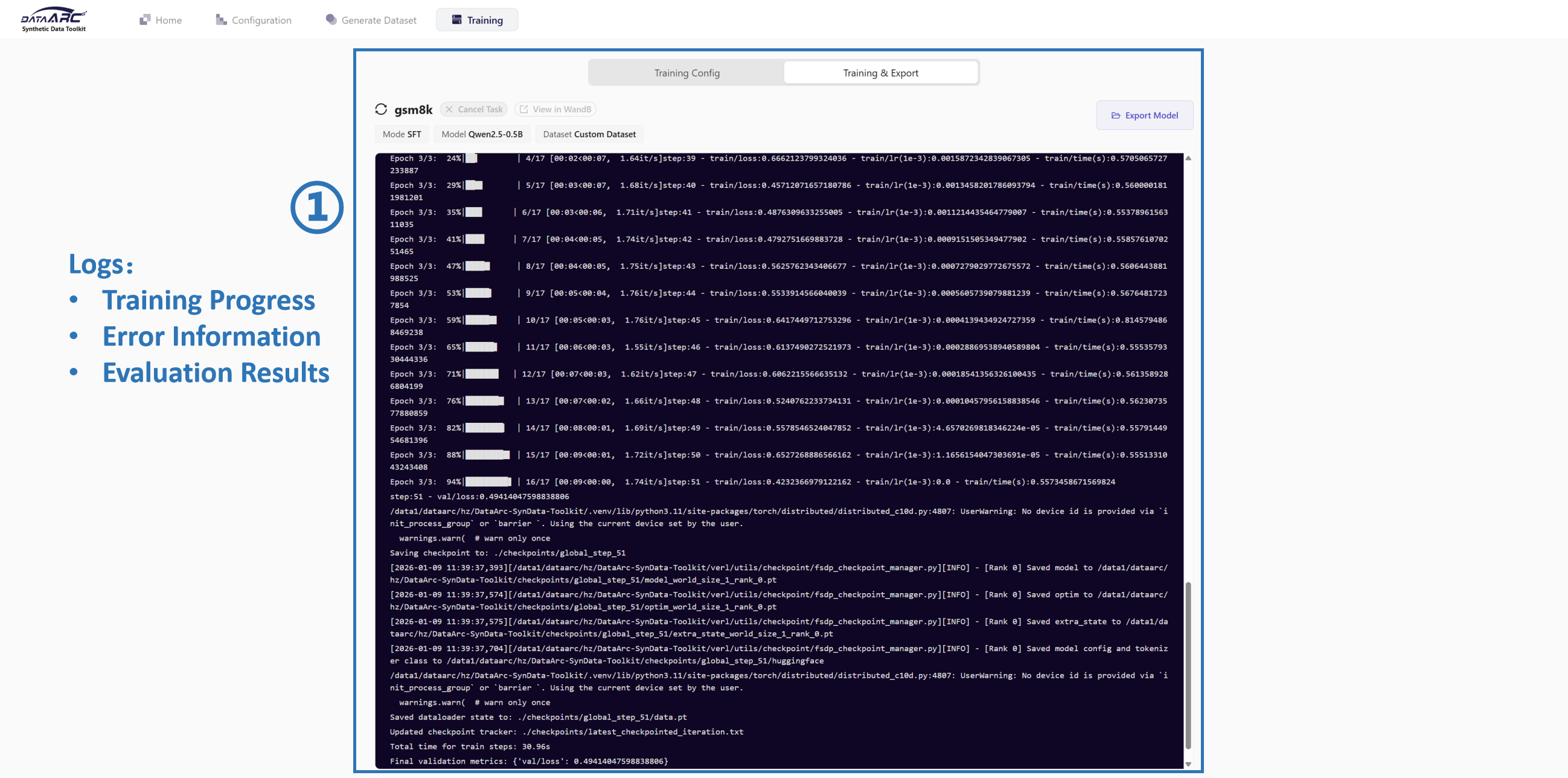}
    \caption{Visualization of training and evaluation.}
    \label{fig:i4_training}
    \vspace{-1em}
\end{figure}

In this section, we show the detailed interface design of DataArc-SynData-Toolkit.
The tutorial video is available on GitHub or on YouTube\footnote{\url{https://www.youtube.com/watch?v=zIHH3YnZKr4&t=56s}}.
The interface design follows the principles of intuitiveness, ease of use, and inspectable process details. 
The toolkit provides interactive interfaces that cover the entire pipeline, including synthetic task configuration (Figure~\ref{fig:i1_task_config}), synthesis visualization (Figure~\ref{fig:i2_data_synthesis}), model training configuration (Figure~\ref{fig:i3_train_config}), and training visualization (Figure~\ref{fig:i4_training}).

As shown in Figure~\ref{fig:i1_task_config}, the configuration interface provides a panel for selecting the synthesis path (Part 1) and a panel for flexible task parameter configuration (Part 2). 
In Part 1, users can choose from Local, Web, or Distill, and upload local corpora (Local), set the HF\_token (Web), or configure a teacher model (Distill). 
In Part 2, users can set task instructions, synthesis quantity, language, and custom prompt templates.
The toolkit provides interaction options with high flexibility.

As shown in Figure~\ref{fig:i2_data_synthesis}, the synthesis interface visualizes the data synthesis, validation, and rewriting steps. 
In Part 1, users can view completed steps and the current execution step. 
In Part 2, users can click on any piece of synthesized data to inspect its details (e.g., input and output) and download them. 
In Part 3, users can monitor the synthesis progress and observe time and token consumption, which allows them to stop the process when needed.

As shown in Figure~\ref{fig:i3_train_config}, the configuration interface is similar to the synthesis task configuration interface. 
It provides a basic configuration panel for uploading models and data (Part 1) and a panel of detailed hyperparameter configurations (Part 2). 
As shown in Figure~\ref{fig:i4_training}, users can directly observe the backend logs, which help users observe training progress, error information, and final evaluation results. 
The visualization of the training process allows users to inspect process details and report issues to developers when errors occur.

Through these visualization interfaces, DataArc-SynData-Toolkit provides interactive support for task configuration, data synthesis, and model training. 
It supports users with different technical backgrounds and offers high flexibility. 
This design significantly improves the usability of the toolkit.
\section{Examples of Synthetic Data}
\label{appendix: sample}

\begin{figure*}[ht]
    \centering
    \includegraphics[width=\linewidth]{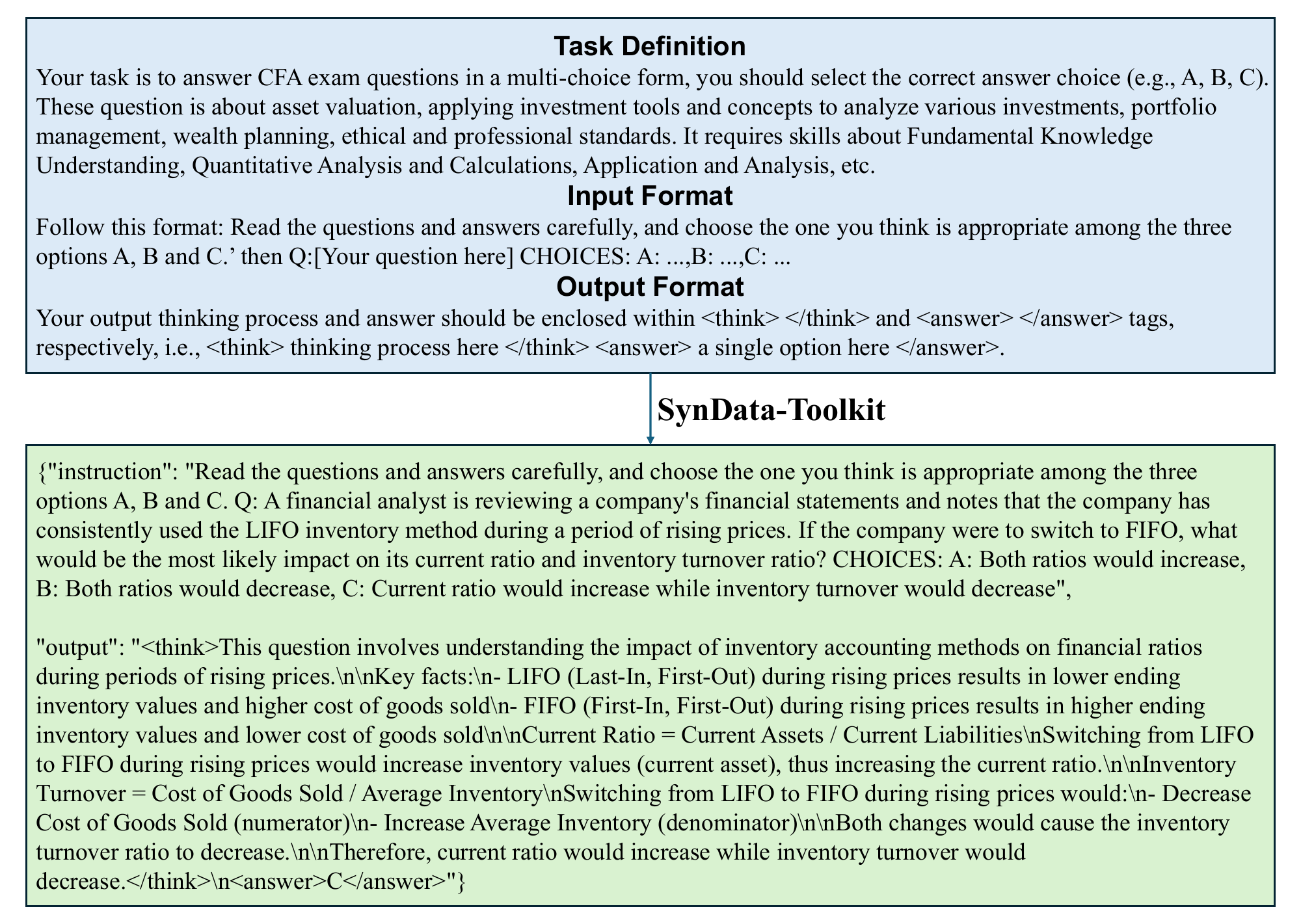}
    \caption{A sample of synthesized instruction in the finance domain.}
    \label{fig:sample_fin}
    \vspace{-1em}
\end{figure*}

\begin{figure*}[ht]
    \centering
    \includegraphics[width=\linewidth]{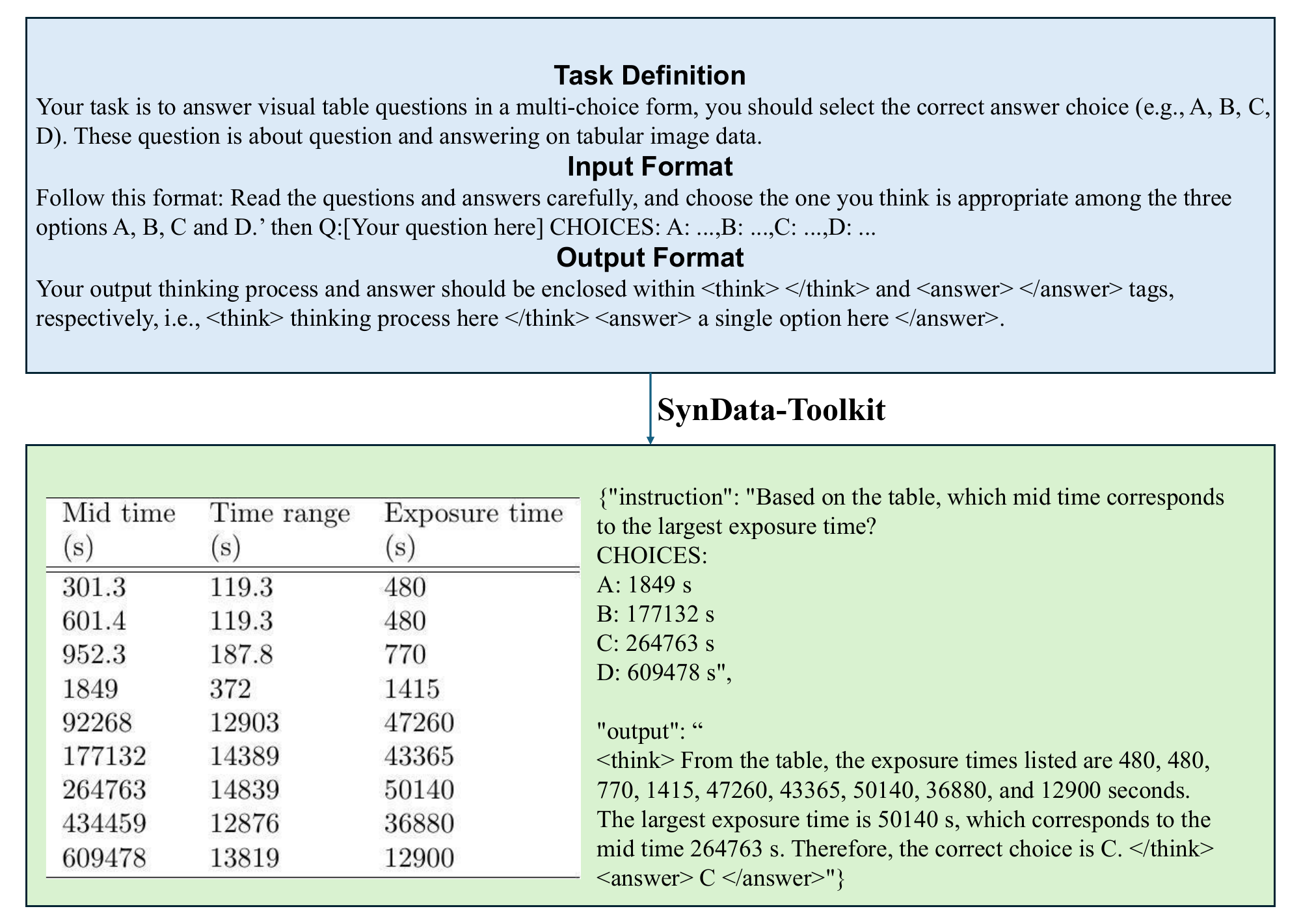}
    \caption{A sample of synthesized instruction in the multimodal domain.}
    \label{fig:sample_mm}
    \vspace{-1em}
\end{figure*}

In this section, we show samples of using our toolkit to synthesize financial data (Figure~\ref{fig:sample_fin}) and multimodal data (Figure~\ref{fig:sample_mm}).

\end{document}